\documentclass{article}
\usepackage{spconf,amsmath,graphicx}
\usepackage{times}
\usepackage{epsfig}
\usepackage{amssymb}
\usepackage{subfigure}
\usepackage{xcolor}
\usepackage{algorithm}
\usepackage{algpseudocode}

\newcommand\eqnref[1]{Eqn.~(\ref{#1})}
\newcommand\figref[1]{Fig.~(\ref{#1})}

\newcommand\tabref[1]{Tab.~\ref{#1}}

\newcommand{\etal}{~et al.}

\title{Shape and Reflectance Reconstruction in Uncontrolled Environments by Differentiable Rendering}
\name{Rui Li, Guangming Zang, Miao Qi, Wolfgang Heidrich}
\address{King Abdullah University of Science and Technology\\\{rui.li, guangming.zang, miao.qi, wolfgang.heidrich\}@kaust.edu.sa}

\begin{document}
%
\maketitle
\begin{abstract}
Simultaneous reconstruction of geometry and reflectance in uncontrolled environments remains a challenging problem. In this paper, we propose an efficient method to reconstruct the scene's 3D geometry and reflectance from multi-view photography using conventional hand-held cameras. Our method automatically builds a virtual scene in a differentiable rendering system that roughly matches the real world's scene parameters, optimized by minimizing photometric objectives alternatingly and stochastically. With the optimal scene parameters evaluated, photo-realistic novel views for various viewing angles and distances can then be generated by our approach. We present the results of captured scenes with complex geometry and various reflection types. Our method also shows superior performance compared to state-of-the-art alternatives in novel view synthesis visually and quantitatively.
\end{abstract}
\begin{keywords}
differentiable rendering, 3D reconstruction, viewpoint synthesis
\end{keywords}

\section{Introduction}\label{sec:intro}
An object's appearance is affected by many internal and external factors, including 3D geometry, surface reflection, and transmission, environmental lighting conditions, viewing angle, or camera position, etc. We call all of these the scene parameters.  
To estimate the scene parameters from multi-view photograph is a emerging and challenging task.
It involves several inter-connected sub-problems, many previous works are focusing on recovering one of those parameters. To name a few: 3D shape reconstruction~\cite{liu2019soft, pix3d, geng2011structured}, image-based rendering~\cite{chan2007image,zhang2004survey,liu2017new}, spatially-varying bidirectional reflectance distribution functions (SVBRDF) acquisition~\cite{li2020inverse, luan2021unified}, environment lighting estimation~\cite{Gardner_2019_ICCV, deeplightingCVPRW2020}, structure from motion~\cite{schoenberger2016sfm}, multi-view stereo~\cite{schoenberger2016mvs}, shading and stereo~\cite{maurer2018combining}, reconstructing textured meshes~\cite{thies2019deferred}, etc.

Different from explicit representation of 3D scene or viewpoint, representing 3D scene by implicit function or neural network (e.g., SRN~\cite{sitzmann2019scene}, NeRF~\cite{mildenhall2020nerf}, pixelNeRF\cite{yu2020pixelnerf}, NSVF~\cite{liu2020neural}, MipNeRF~\cite{barron2021mipnerf}) has gained popularity recently by a implicit volume (neural network), and optimizes scene volume via training a self-supervised neural network. 

Differential rendering becomes another promising categories for 3D geometry optimization, texture mapping, novel view rendering. Liu\etal~\cite{liu2019soft} propose a differentiable rendering framework that directly forward render colorized mesh and back-propagate gradient to mesh vertices, color, silhouette, etc. 
Nimier-David\etal~\cite{NimierDavidVicini2019Mitsuba2} proposed a physically-based engine, Mitsuba2, for estimating parameter gradient that based on Monte-Carlo sampling, thus, enables back propagating geometry, and other general scene parameters gradient flow. 

However, directly optimizing scene parameters especially geometry structure of wild scene is still a challenging problem, experiments show the limited toy cases for directly optimization of geometry.
Specifically, there are several technical
challenges: first, accurate 3D geometry is unknown or requires an
expensive 3D scanner. Second, specular reflection is shape-sensitive
having a significant influence on object appearance. Finally, the
natural environment contains multiple direct light sources and an
indirect ray path, which is unknown and hard to direct.

%
%



To overcome the above technical challenges and enable reconstruction
in a typical user scenario, we propose a systematic scene parameter
reconstruction method that jointly estimates 3D geometry, surface
reflectance, specular coefficients, camera pose, position, and
lighting condition using a differentiable inverse rendering
framework. Note that there are no specific requirements for the
the photography acquisition process, i.e., it only requires several
multi-view photos or surround video, without extra constraints such as
controlled lighting, camera position or exposure, or a specific environment setup, to enable in-the-wild reconstruction.

Our contributions are listed as follows, (1) we propose a general parameterized framework to describe typical object appearance, enabling the direct reconstruction of a realistic scene from real-world multi-view photography for uncontrolled lighting conditions and general diffuse and specular scene. Thus, it can work entirely in the wild. (2) We propose a memory-efficient solution for differentiable forward rendering and backward propagation.  Our framework can directly optimize real-world scene parameters in an iterative and stochastic fashion. (3) Our method can also enable several photo-realistic applications such as novel view synthesis, environment light editing, etc.

\section{Proposed Method}\label{sec:method}
\begin{figure}[ht]
	\centering
	\includegraphics[width=0.99\linewidth]{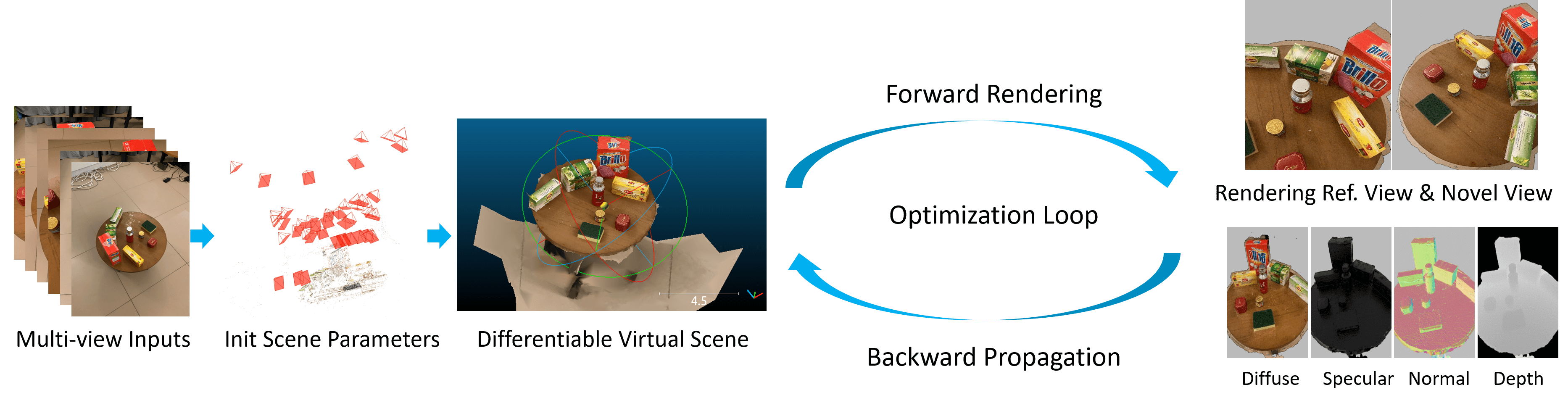}
	\caption[]{Inverse parameter optimization pipeline.}
	\label{fig:tbd}
\end{figure}

We illustrate our proposed framework in \figref{fig:tbd}: It takes a set of RGB images $I = \{I_k\}$ of the scene from arbitrary viewpoints and camera position as input. Then, our method builds a virtual scene that roughly matches real world camera poses, and iteratively optimize interested scene parameters for 3D geometry, diffuse and specular reflectance, and environment lighting by pushing the photometric consistency between rendering images and real photography. Finally, a set of optimized scene parameters can enable several photo-realistic applications: view synthesis, lighting editing, etc.
Mathematically, our system can be described as a function of desired parameters as, 
\begin{equation}\label{eqn:image_formation}
I = \Phi(\theta_{g}, \theta_{d}, \theta_{s}, \theta_{l}),
\end{equation} 
where $\Phi$ is the physically-based rendering process, which
simulates the light rays traveling in a virtual scene.  $\theta_{g} =
\{x_0, \dots, x_N\}$ is the set of 3D positions corresponding to the
mesh vertices.  $\theta_{d} = \{\theta_{d}(x_0), \dots,
\theta_{d}(x_N)\}$ is the set of diffuse per-vertex color
reflectance, with $\theta_{d}(x) \in \mathbb{R}^{3}$.  Similarly, $\theta_{s}$
is the set of specular reflectances.
The illumination in the scene is modeled as environmental lighting $l_0$,
approximated as a set of isotropic point sources $\theta_{l} = \{l_0, l_1, \dots, l_M\}$, $l_n$ is the intensity for $n$-th light sources.  
We use
Mitsuba2~\cite{NimierDavidVicini2019Mitsuba2} as our physically-based
rendering engine and renderer. Therefore,
it enables an iterative optimization pipeline that back propagates
rendering error to update scene parameters.

Our image formation model for each 2D pixel coordinate $u$ can be formulated as 
\begin{equation}\label{eqn:image}
    I(u) = L(w_o; x)\Delta{t},
\end{equation}
where $x \in \mathbb{R}^{3}$ is the 3D coordinate, and $u \in
\mathbb{R}^{2}$ is the 2D image coordinate. $L(w_o; x)$ is the
radiance from reflected scene point $x$ with outgoing direction
$w_o$. $\Delta{t}$ is the exposure time.  According to rendering
equation~\cite{RenderingTOG1986}, the radiance of a non-emitting
object can be written as
\begin{equation}\label{eqn:reflectance}
	L_{o}(w_o; x) = \int_{\Omega} f_r(x, w_i, w_o) L_{i}(w_i; x) (w_i \cdot n(x)) \,\mathrm{d} w_i, 
\end{equation}
where $L_{i}(w_i; x)$ and $L_{o}(w_o; x)$ are the incoming and outgoing radiance functions with direction $w_o$ and $w_i$ for a 3D physical point $x$.
$n(x)$ is the normal function. $\Omega$ is the ray direction space. $f_r$ is physical or estimated reflectance model.

In our virtual scene, an environment light source emits light to
the virtual scene and bounces when hitting a 3D object. This
environment light is represented as a set of isotropic point sources.
Let $x^{\prime}$ be the 3D position of one of the environment light
point sources.
The ray tracing can be described by an iterative process. First, light
rays originate at an environment light source as

\begin{equation}\label{eqn:ray_trace1}
L^{0}(w_i; x) = \theta_{l}(x^{\prime}),
\end{equation}
where the superscript notes the 0-{\em th} bounce of light.
We assume that environment light is an isotropic light. Thus, the intensity is identical for any incoming direction $w_i$, i.e., $x^{\prime} \rightarrow x$. 
When the light ray hitting the 3D object in the scene, 
$L^{t}_{i}(w_o; x)$ is the $t$-{\em th} bounce of outgoing light ray
that directly hits camera aperture, and other reflected light with a
different direction than $w_o$ will start a new bounce until reaching
a maximal bounce limit.
\begin{equation}\label{eqn:ray_trace2}
	L^{t}_{o}(w_o; x) = \int_{\Omega} f_r(x, w_i, w_o) L^{t-1}_{i}(w_i; x) (w_i \cdot n(x)) \,\mathrm{d} w_i, 
\end{equation}
the overall  radiance received by the camera is the sum over all
outgoing light, and a maximum of $T$ bounces:
\begin{equation}\label{eqn:ray_trace3}
	L_{o}(w_o; x) = \sum_{t=0}^{T} L^{t}_{o}(w_o; x),
\end{equation}

We assume that our scene contains a rough surface with diffuse and
specular reflection without transmission, and Cook-Torrance (CT)
model~\cite{BeckmanTOG1982} with an optional microfacet distribution
function, e.g., Beckmann~\cite{BeckmanTOG1982}, GGX~\cite{GGXEG2007},
can describe a broad class of general real-world objects in reflection. 
Our reflectance model $f_r$ can be expressed as follows:

\begin{eqnarray}\label{eqn:diffuse_specular_ref}
f_r(x, w_{i}, w_{o}) &=& \theta_{d}(x) + \theta_{s}(x, w_{i}, w_{o}), \\
\theta_{d}(x)&=&\frac{\rho_{d}(x)}{\pi},\\
\theta_{s}(x, w_{i}, w_{o})&=&\rho_{s}(x) \frac{D(h, \alpha) G(n(x), w_{i}, w_{o}) F(h, w_{i})}{4(n(x) \cdot w_{i})(n(x) \cdot w_{o})},
\end{eqnarray}
where $\theta_{d}$ and $\theta_{s}$ are diffuse and specular reflectance respectively, $\rho_{d}$ and $\rho_{s}$ are diffuse and specular albedos, $h$ is the halfway vector, which is computed by normalizing the sum of the light direction $w_i$ and view direction vectors $w_o$. 
Our $D(h, \alpha)$ is the microfacet distribution function. 
$\alpha$ specifies the roughness of surface micro-geometry along with the tangent and bitangent directions.
$G$ is a shadowing-masking function, and $F$ is the Fresnel term similar in \cite{GGXEG2007}. 
Thus, we reach our rendering function by combining \eqnref{eqn:image}, \eqnref{eqn:reflectance}, \eqnref{eqn:diffuse_specular_ref} as,
\begin{equation}\label{eqn:rendering_func}
    I(u) = \Phi(\theta_{g}, \theta_{d}, \theta_{s}, \theta_{l})(x) = \Delta{t}\sum_{t=0}^{T} L^{t}_{o}(w_o; x),
\end{equation} 

\begin{equation}\label{eqn:rendering_func2}
	I(u) = \Delta{t}\sum_{t=0}^{T} \int_{\Omega} (\theta_{d}(x)+\theta_{s}(x)) L^{t-1}_{i}(w_i; x) (w_i \cdot n(x)) \,\mathrm{d} w_i,
\end{equation}

$L^{t-1}_{i}(w_i; x)$ is incoming radiance from every possible direction with a bounce number of $t-1$, which is also an integral over the bounce number of $t-2$. For the case of $t=0$ in \eqnref{eqn:ray_trace1}, $L^{0}(w_i; x)$ is initial environment lighting $\theta_{l}$.  

Our objective function aims at jointly reconstructing the diffuse reflectance $\theta_{d}$ and specular reflectance $\theta_{s}$, 3D geometry $\theta_{g}$ for mesh and environment light source $\theta_l$.

\begin{equation}\label{eqn:obj_all}
\mathcal{O} = \sum_{k=1}^{K} \|M_{k}I_{k} - \Phi_{k}(\theta_{g}, \theta_{d}, \theta_{s}, \theta_{l})\|_{2}^{2}.
\end{equation}
Rendering results only
contain the target object without background, and thus the error of
the background pixels will dominate objective value.
To alleviate this effect, $M_k$ is a pre-computed binary mask that removes background pixel contribution in the objective calculation, generated by segmentation methods GrabCut~\cite{rother2004grabcut} and ~\cite{lirsiggraphasia2019} for view consistency. 
We implement our differentiable optimization pipeline in Mitsuba2~\cite{NimierDavidVicini2019Mitsuba2} by using forward rendering and backward propagation manner and initial parameters by colmap.
Due to the limited GPU memory, we alternate
between updating each parameter while keeping the others unchanged.
By choosing one scene parameters $\theta \in \{\theta_d, \theta_g, \theta_s, \theta_l\}$, we iteratively update chosen $\theta$ in the inner loop. 

\section{Experiments and Comparison}\label{sec:exp}



We take photo for the target scene from multiple viewpoints (20-40) with an auto-focus setup, without extra controlled lighting or flash, ordinary ambient light or direct, diffuse light is sufficient. At the initial stage, we set the radiance of the light source to be 0.5.
Physically-based rendering is a GPU memory-consuming task, and we adjust several parameters for the sake of memory saving. We set the maximum number of bounce $T=3$ for ray tracing bounce number, and raw images are downsampled for $8\times$, the number of sampling per pixel $spp=1$ in iterative optimization stage, and $spp=16$ for final output with better visual quality. We use Adam as our optimizer, as $\lambda_{d}=0.1$, $\lambda_{g}=0.5$, $\lambda_{s}=0.01$, $\lambda_{l}=0.05$ as learning rate. $\alpha=0.1$ for general surface roughness. Our proposed method runs around $140 ms$ per iteration, 400 iterations to optimize a viewpoint.

\subsection{Evaluation and comparison}
\begin{figure*}[htbp]
	\centering
	\def \scale {0.1}
	\includegraphics[width=\scale\linewidth]{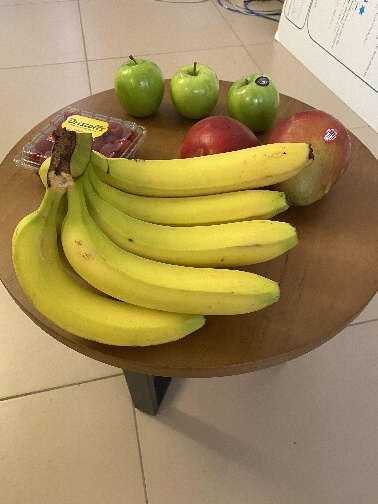}
	\includegraphics[width=\scale\linewidth]{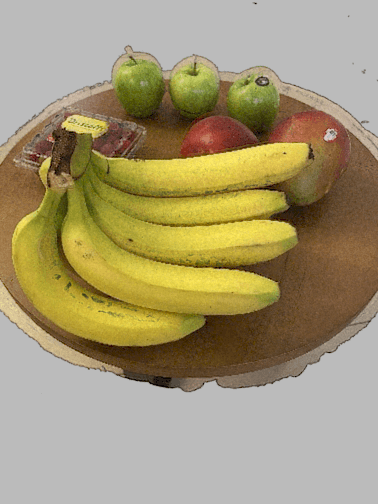}
	\includegraphics[width=\scale\linewidth]{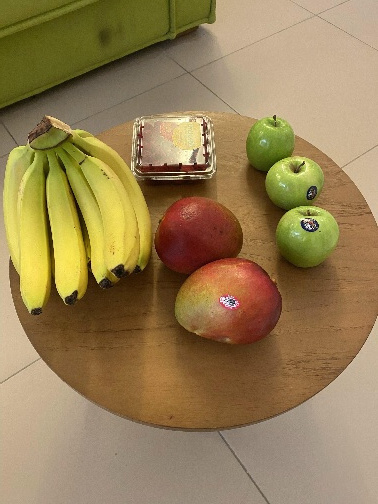}
	\includegraphics[width=\scale\linewidth]{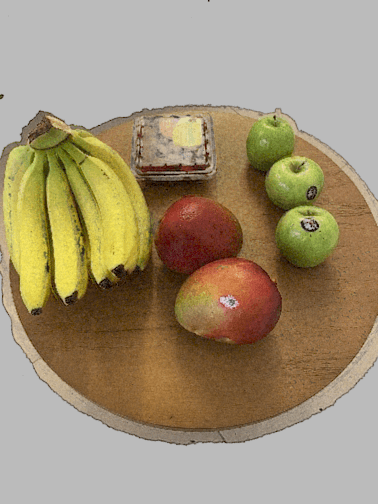}
	\includegraphics[width=\scale\linewidth]{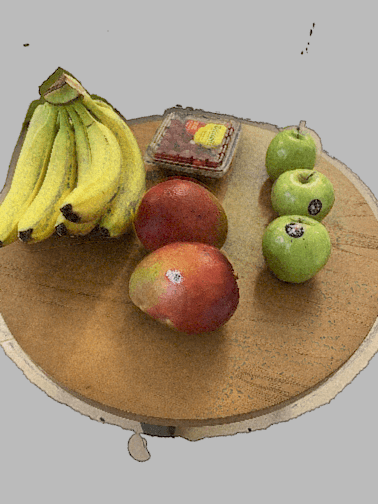}
	\includegraphics[width=\scale\linewidth]{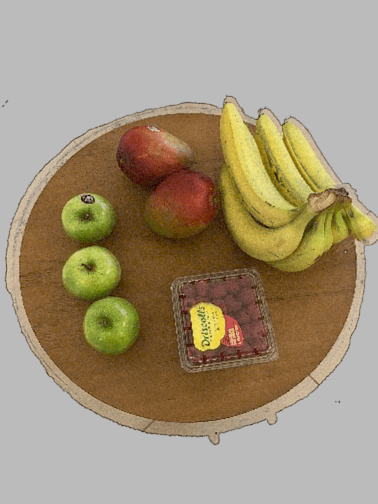}
	\includegraphics[width=\scale\linewidth]{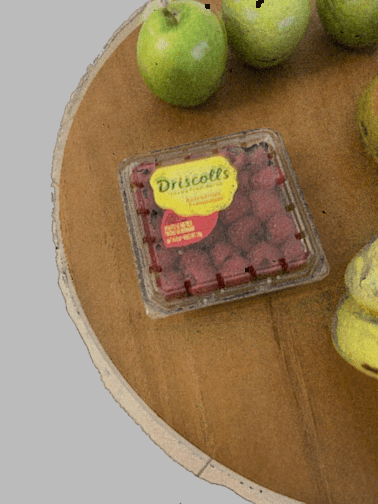}
	\includegraphics[width=\scale\linewidth]{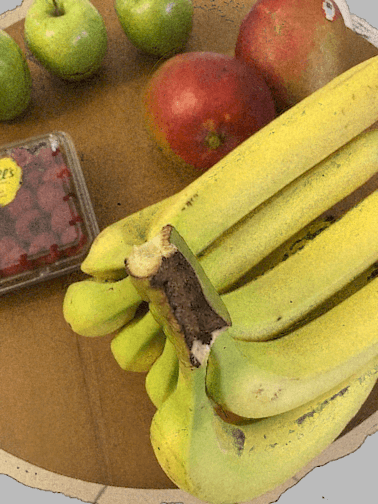}\\
	
	\includegraphics[width=\scale\linewidth]{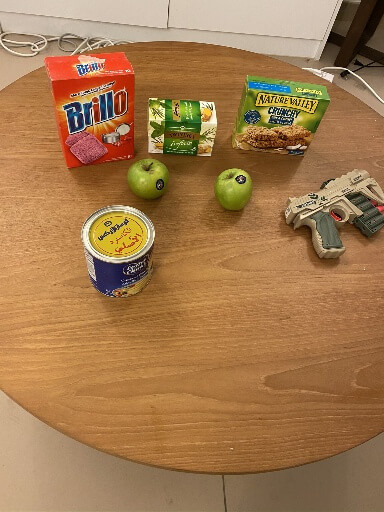}
	\includegraphics[width=\scale\linewidth]{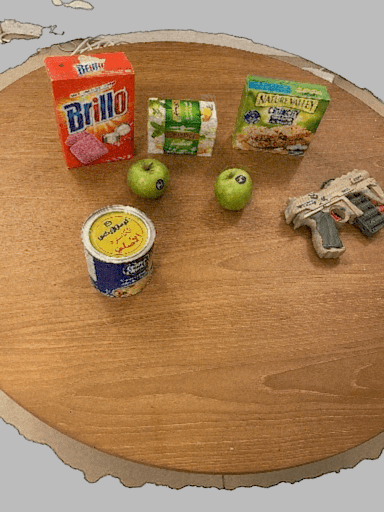}
	\includegraphics[width=\scale\linewidth]{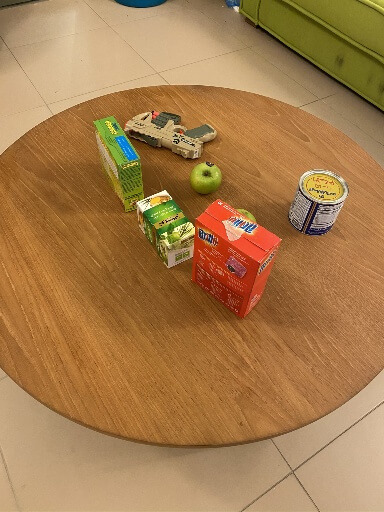}
	\includegraphics[width=\scale\linewidth]{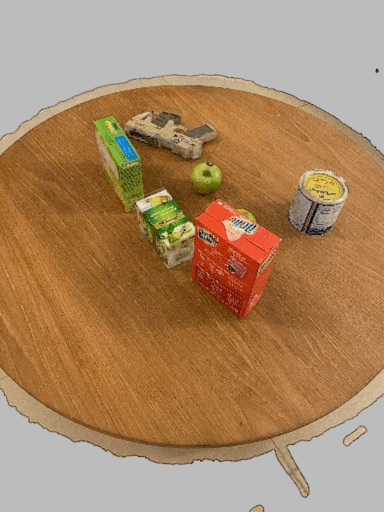}
	\includegraphics[width=\scale\linewidth]{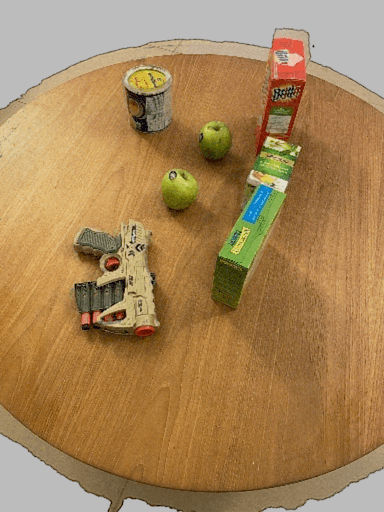}
	\includegraphics[width=\scale\linewidth]{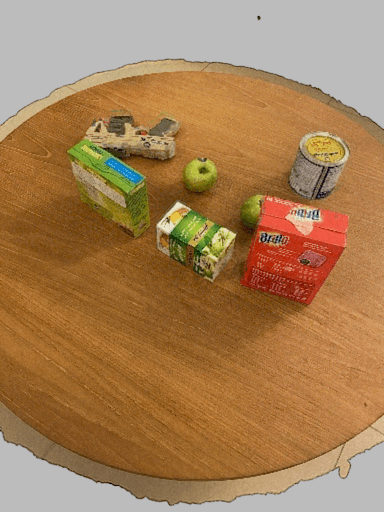}
	\includegraphics[width=\scale\linewidth]{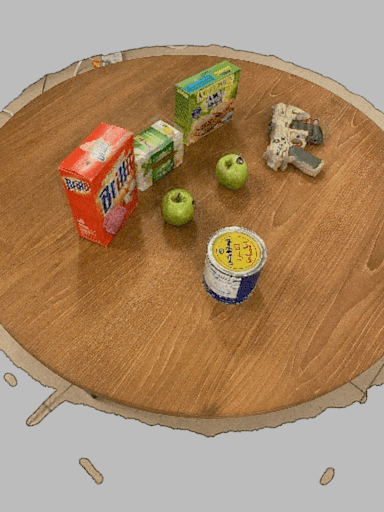}
	\includegraphics[width=\scale\linewidth]{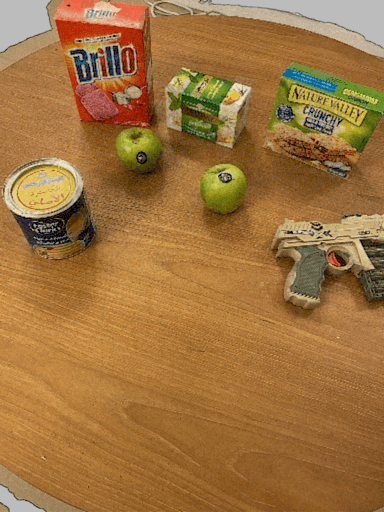}
	\caption{Full scene reconstruction. Two scenes are shown: \texttt{Fruit} (Row 1) and \texttt{Table 2} (Row 2). Column 1, 3 is real captured images, column 2, 4 is corresponding virtual view, column 5-8 are rendering synthetic novel views.}
	\label{fig:rec2}
\end{figure*}

\begin{figure}[htbp]
	\centering
	\def \scale {0.15}
	\includegraphics[width=\scale\linewidth]{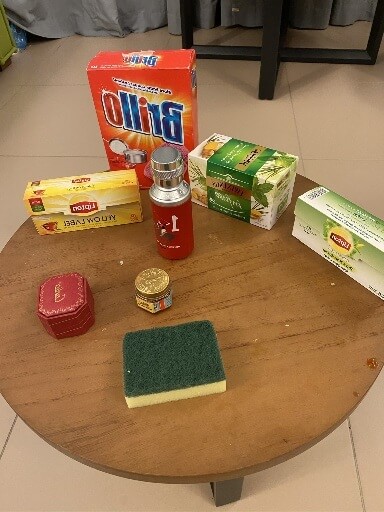}
	\includegraphics[width=\scale\linewidth]{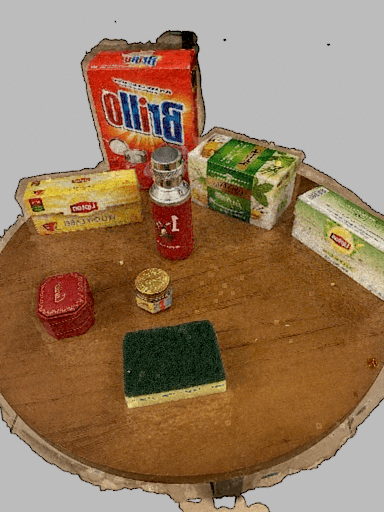}
	\includegraphics[width=\scale\linewidth]{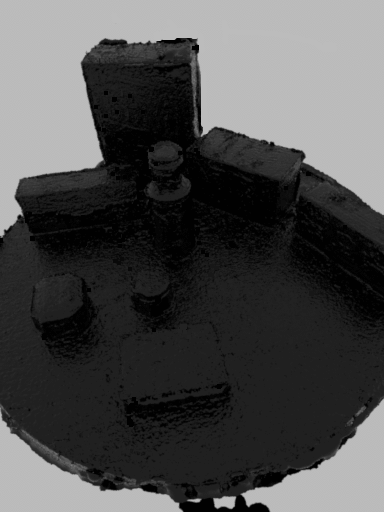}
	\includegraphics[width=\scale\linewidth]{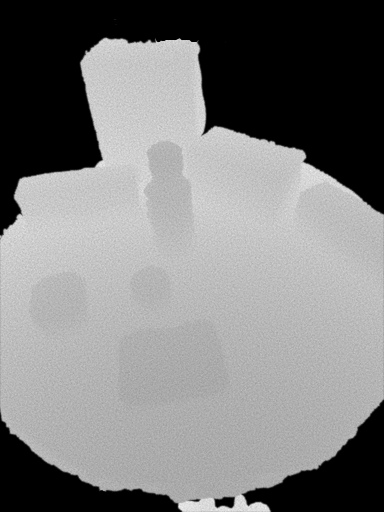}
	\includegraphics[width=\scale\linewidth]{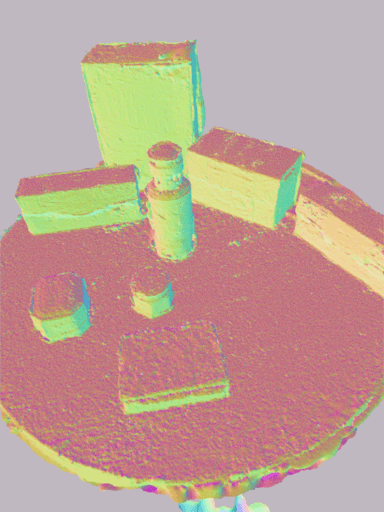}
	\includegraphics[width=\scale\linewidth]{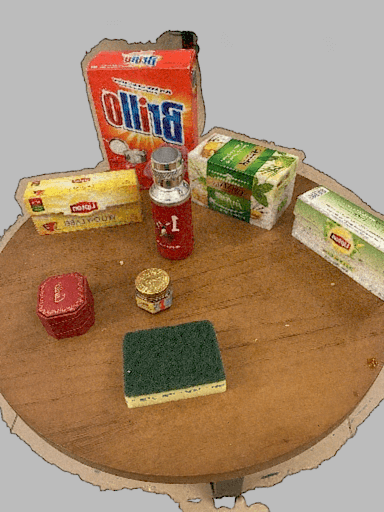}\\

	\caption{Reconstruction Components. From left to right: real scene photography, diffuse reflectance $\theta_{d}$, specular reflectance $\theta_{s}$, depth map, shading normal, rendering scene. 
	}
	\label{fig:rec_result}
\end{figure}

In this section, we compare our method against other SOTA methods visually and quantitatively.
\figref{fig:rec2} shows reconstructed scenes and novel view rendering. Our novel view rendering can successfully recover true 3D geometry of the scene, accurate texture and details of objects, the photo-realistic glossy reflection of the surface. 
\figref{fig:rec_result} shows the recovered components of diffuse and specular reflectance, novel views, generated depth, and surface normal.


\begin{figure}[h]
	\centering
	\includegraphics[width=0.9\linewidth]{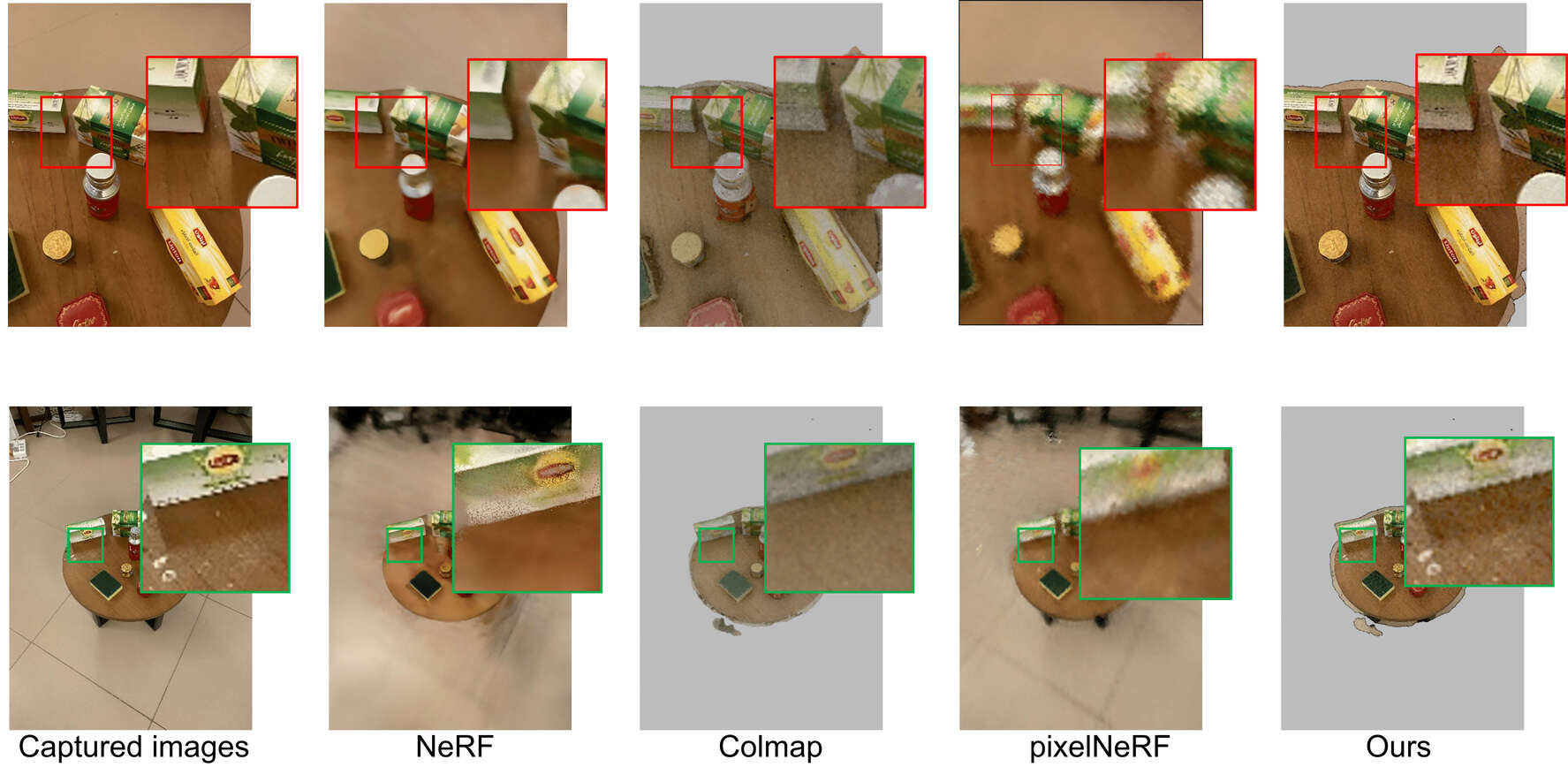}
	\caption{Viewpoint synthesis for various camera poses and range. From left to right: captured images (\texttt{Table 1}), NeRF results~\cite{mildenhall2020nerf}, Colmap~\cite{schoenberger2016mvs}, pixelNeRF~\cite{yu2020pixelnerf} and Ours.}
	\label{fig:comp_view_synthesis}
\end{figure}

For visual comparison in \figref{fig:comp_view_synthesis}, our input images contain significant viewpoint changes and various captured distances, NeRF~\cite{mildenhall2020nerf} have leaking light phenomenon in a long range view, due to less novel view direction sampling. We render Colmap~\cite{schoenberger2016sfm,schoenberger2016mvs}'s output textured mesh in renderer, thus, as anticipated, rendering results with decent 3D structures but less accurate surface reflectance can be acquired.  PixelNeRF~\cite{yu2020pixelnerf} successfully reconstructs 3D geometry with various camera distances using a pretrained network, but still has poor performance in recovering detail textures. NeRF-based methods apply implicit volume to describe diffuse/specular reflection real scene, without taking multiple bounce of light and geometry cues for solving inverse rendering problem.
In contrast, 
our differentiable pipeline directly optimizes the scene parameter to match the real scene image. Therefore, it can accurately achieve photo-realistic high-quality performance in these viewpoint synthesis scenarios.

\begin{figure}[htbp]
	\def \scale {0.8}
	\centering
	\includegraphics[width=\scale\linewidth]{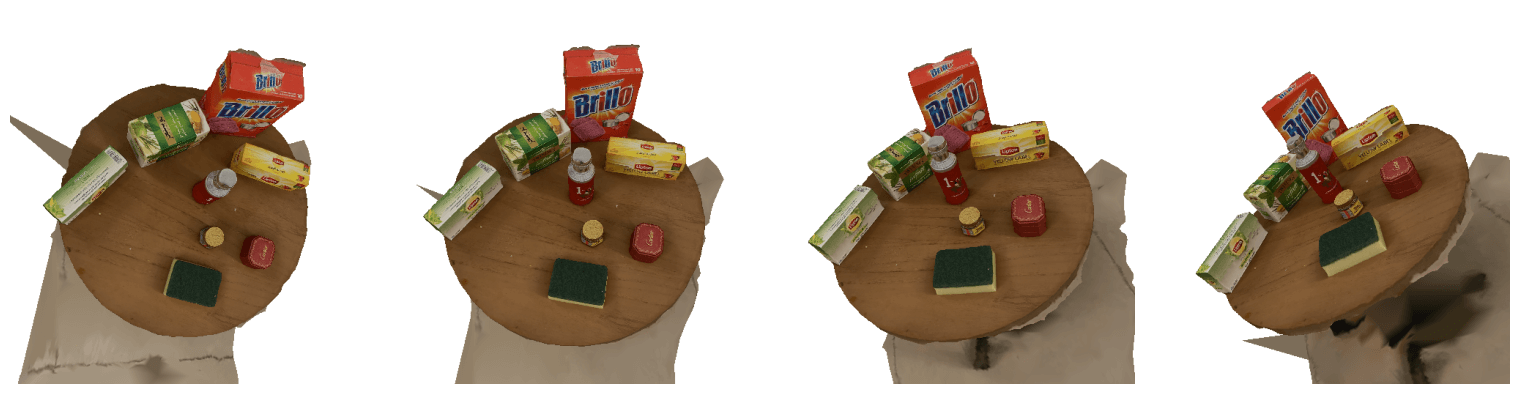}\\
	\includegraphics[width=\scale\linewidth]{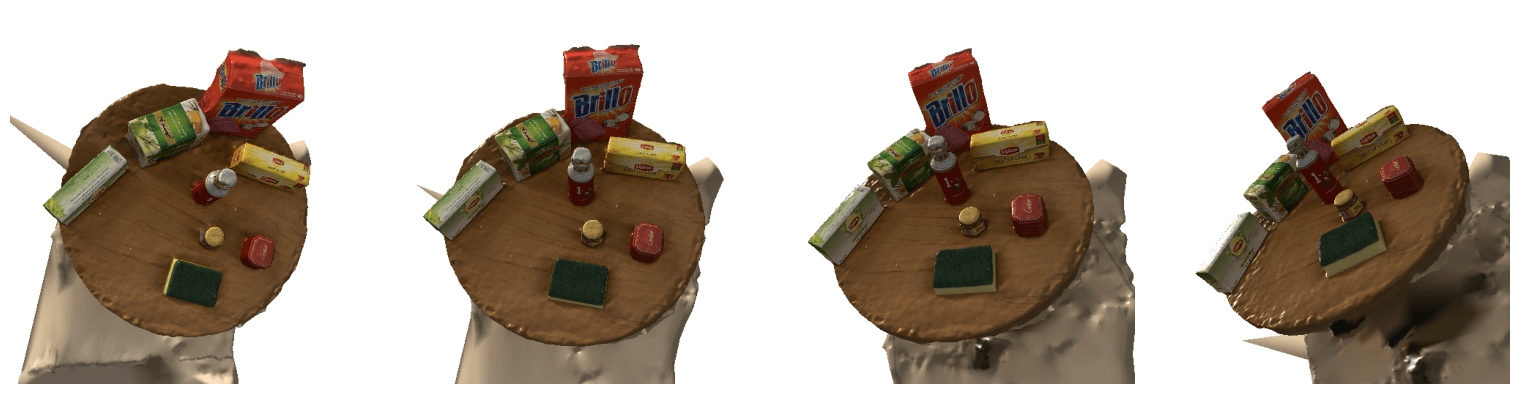}\\
	\includegraphics[width=\scale\linewidth]{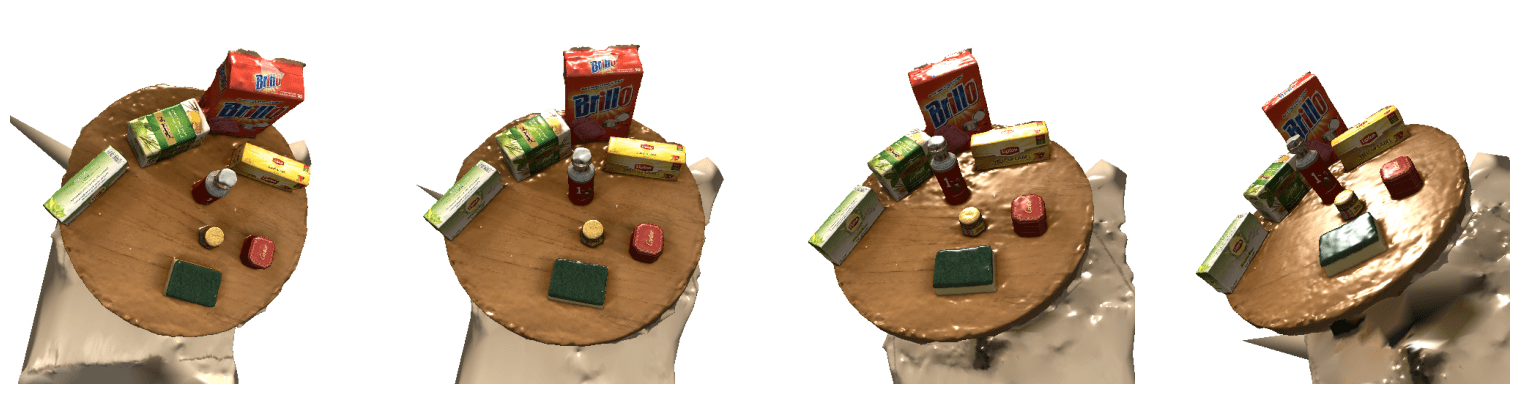}
	\caption{Relighting. R1: point light source with fixed position. R2: directional + ambient light. R3: various directional light.}
	\label{fig:relighting}
\end{figure}

\begin{figure}[htbp]
	\centering
	\def \scale {0.15}
	\includegraphics[width=0.27\linewidth]{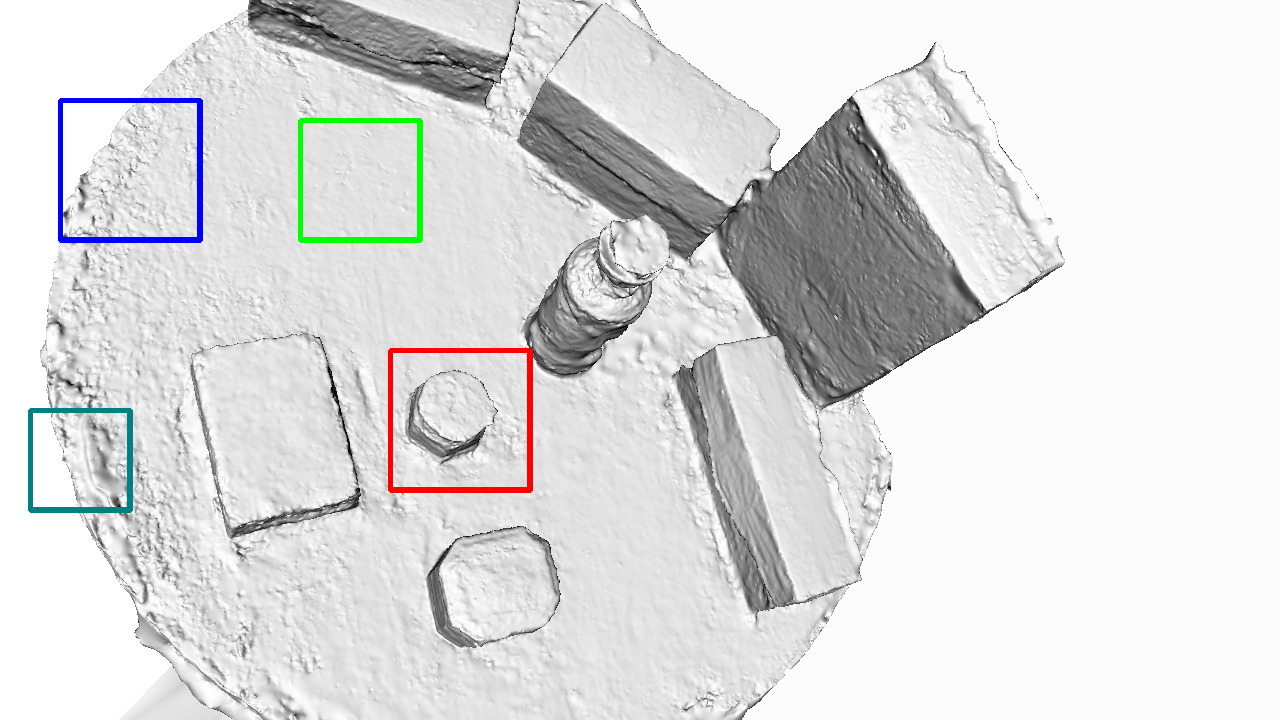}
	\includegraphics[width=\scale\linewidth]{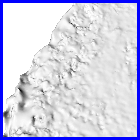}
	\includegraphics[width=\scale\linewidth]{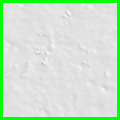}
	\includegraphics[width=\scale\linewidth]{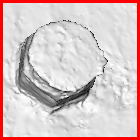}
	\includegraphics[width=\scale\linewidth]{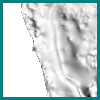}\\
	\includegraphics[width=0.27\linewidth]{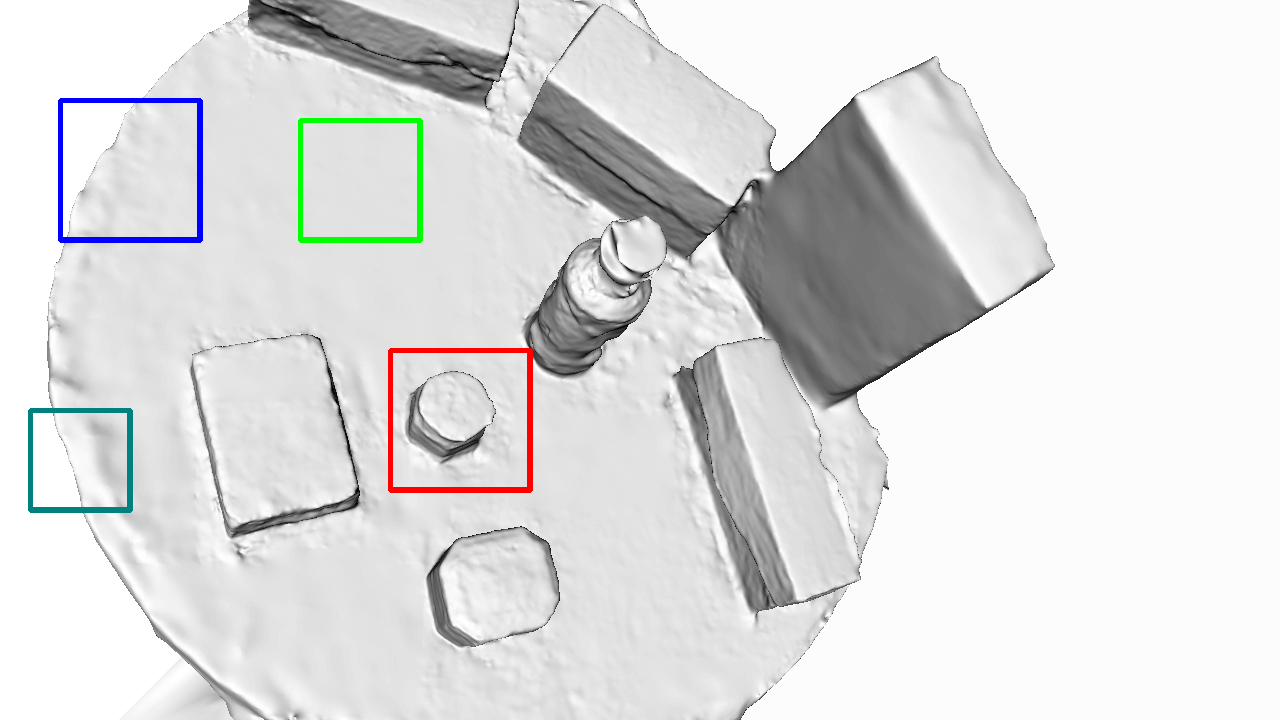}
	\includegraphics[width=\scale\linewidth]{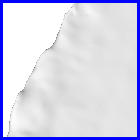}
	\includegraphics[width=\scale\linewidth]{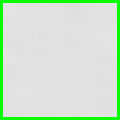}
	\includegraphics[width=\scale\linewidth]{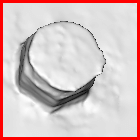}
	\includegraphics[width=\scale\linewidth]{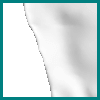}\\
	\caption{Surface Refinement. R1: original geometry, R2: optimized geometry. Our method can significantly correct local geometry error in feature matching.}
	\label{fig:surface_rec}
\end{figure}

\begin{figure}[h]
	\def \scale {0.18}
	\def \scaleB {1.0}
	\centering
	\subfigure[Real]{
		\begin{minipage}[t]{\scale\linewidth}
			\includegraphics[width=\scaleB\linewidth]{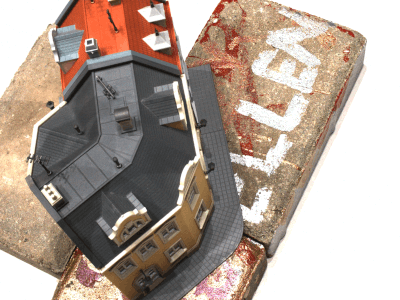}
			\includegraphics[width=\scaleB\linewidth]{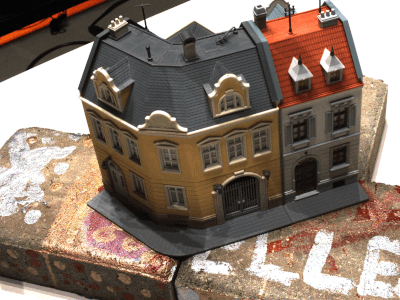}
	\end{minipage}}
	\subfigure[NeRF]{
		\begin{minipage}[t]{\scale\linewidth}
			\includegraphics[width=\scaleB\linewidth]{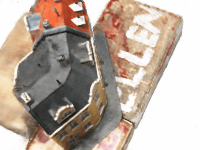}
			\includegraphics[width=\scaleB\linewidth]{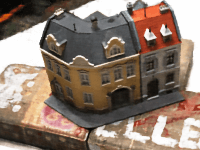}
	\end{minipage}}
	\subfigure[colmap]{
		\begin{minipage}[t]{\scale\linewidth}
			\includegraphics[width=\scaleB\linewidth]{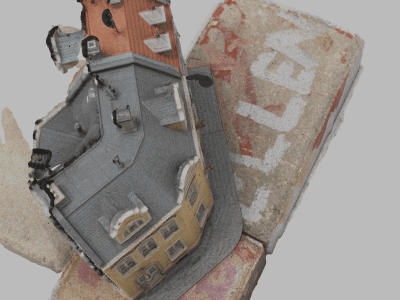}
			\includegraphics[width=\scaleB\linewidth]{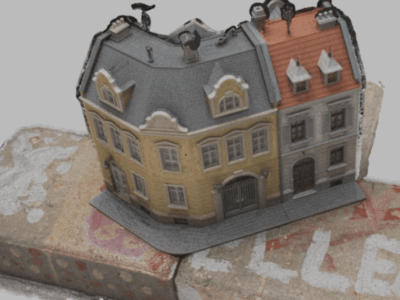}
	\end{minipage}}
	\subfigure[pNeRF]{
		\begin{minipage}[t]{\scale\linewidth}
			\includegraphics[width=\scaleB\linewidth]{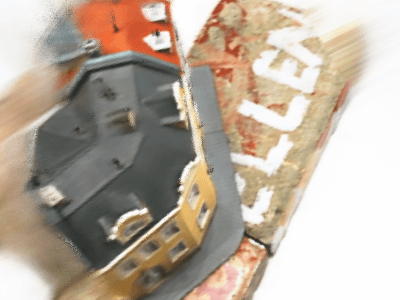}
			\includegraphics[width=\scaleB\linewidth]{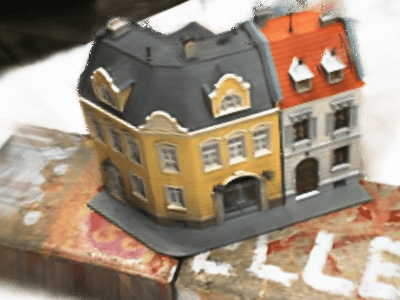}
	\end{minipage}}
	\subfigure[Our]{
		\begin{minipage}[t]{\scale\linewidth}
			\includegraphics[width=\scaleB\linewidth]{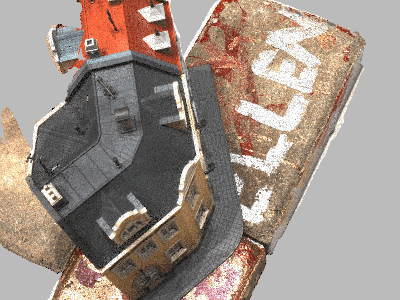}
			\includegraphics[width=\scaleB\linewidth]{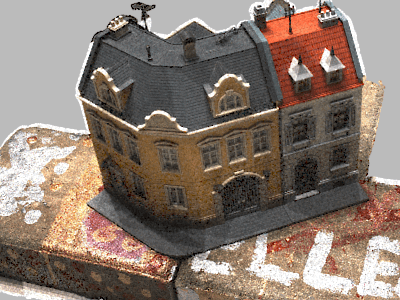}
	\end{minipage}}
	\caption{Visual comparison on DTU\cite{jensen2014large} dataset.}
	\label{fig:compare_dtu}
\end{figure}

In \figref{fig:compare_dtu}, we test our method with other SOTA methods by using public dataset DTU.
As anticipated, colmap~\cite{schoenberger2016sfm} show inaccurate texture recovery, pixelNeRF~\cite{yu2020pixelnerf} will blur the rendering images with inaccurate lighting since it only use several views as inputs. NeRF~\cite{mildenhall2020nerf} successfully recover near view scene in DTU cases, however, slightly blur the synthetic view, while our method performs better in details recovery. In \figref{fig:surface_rec}, we visualize the effect of our pipeline for optimizing surface and correct geometry error by utilizing multi-view inverse rendering.
We also show relighting results by adding several type of light sources (point light, directional light, diffuse light) with different light positions and viewing angles in \figref{fig:relighting}.
The quantitative evaluation with PSNR and SSIM measurements between synthetic novel view and the captured image is shown in \tabref{tab:quan}.

\begin{table}[htbp]
	\centering
	\begin{tabular}{|l|cc|cc|}
		\hline
		\multicolumn{1}{|c|}{} & \multicolumn{2}{c|}{\texttt{Fruit}}                           & \multicolumn{2}{c|}{\texttt{Table 1}}                          \\ \cline{2-5} 
		\multicolumn{1}{|c|}{} & \multicolumn{1}{c}{PSNR$\uparrow$} & \multicolumn{1}{c|}{SSIM$
		\uparrow$} & \multicolumn{1}{c}{PSNR$\uparrow$} & \multicolumn{1}{c|}{SSIM$\uparrow$} \\ \hline
		NeRF                   & 14.35                    & 0.28                      & 15.65                    & 0.32                      \\
		PixelNeRF              & 26.25                    & 0.91                      & 26.35                    & 0.84                      \\
		Colmap                 & 17.51                    & 0.70                      & 18.67                    & 0.75                      \\ \hline
		Initial                & 17.51                    & 0.70                      & 18.67                    & 0.75                      \\
		Diff Opt.              & 26.31                    & 0.82                      & 27.21                    & 0.85                      \\
		Geo Opt.               & 27.42                    & 0.93                      & 27.45                    & 0.86                      \\
		Spe Opt.               & 27.45                    & 0.94                      & 27.95                    & 0.87                      \\
		Light Opt.             & 28.78                    & 0.95                      & 28.03                    & 0.87                      \\ \hline
		Ours                   & 29.02                    & 0.96                      & 28.45                    & 0.88                      \\ \hline
	\end{tabular}
	\caption{PSNR\&SSIM evaluations for comparison.}
	\label{tab:quan}
\end{table}

\section{Conclusion}\label{sec:con}
We propose a novel differentiable optimization framework that simultaneously reconstructs scene parameters: diffuse and specular reflectance, geometry, environment lighting using the hand-held camera from an uncontrolled environment. 
Our method can handle a wide range of materials and general ambient lighting, offers an attractive and efficient solution, facilitating in-the-wild scene reconstruction for a wider public, enables various applications: photo-realistic view synthesis, 3D geometry optimization, depth estimation, lighting editing. 



\bibliographystyle{IEEEbib}
\bibliography{trans}

\begin{thebibliography}{10}

\bibitem{liu2019soft}
Shichen Liu, Tianye Li, Weikai Chen, and Hao Li,
\newblock ``Soft rasterizer: A differentiable renderer for image-based 3d
  reasoning,''
\newblock in {\em Proc. ICCV}, 2019.

\bibitem{pix3d}
Xingyuan Sun, Jiajun Wu, Xiuming Zhang, Zhoutong Zhang, Chengkai Zhang, Tianfan
  Xue, Joshua Tenenbaum, and William Freeman,
\newblock ``Pix3d: Dataset and methods for single-image 3d shape modeling,''
\newblock in {\em Proc. CVPR}, 2018.

\bibitem{geng2011structured}
Jason Geng,
\newblock ``Structured-light 3d surface imaging: a tutorial,''
\newblock {\em Advances in Optics and Photonics}, 2011.

\bibitem{chan2007image}
SC~Chan, Heung-Yeung Shum, and King-To Ng,
\newblock ``Image-based rendering and synthesis,''
\newblock {\em SPM}, 2007.

\bibitem{zhang2004survey}
Cha Zhang and Tsuhan Chen,
\newblock ``A survey on image-based rendering—representation, sampling and
  compression,''
\newblock {\em Signal Processing: Image Communication}, 2004.

\bibitem{liu2017new}
Zhong Liu, Zhouchi Lin, Xiguang Wei, and Shing-Chow Chan,
\newblock ``A new model-based method for multi-view human body tracking and its
  application to view transfer in image-based rendering,''
\newblock {\em TMM}, 2017.

\bibitem{li2020inverse}
Zhengqin Li, Mohammad Shafiei, Ravi Ramamoorthi, Kalyan Sunkavalli, and
  Manmohan Chandraker,
\newblock ``Inverse rendering for complex indoor scenes: Shape,
  spatially-varying lighting and svbrdf from a single image,''
\newblock in {\em Proc. CVPR}, 2020.

\bibitem{luan2021unified}
Fujun Luan, Shuang Zhao, Kavita Bala, and Zhao Dong,
\newblock ``Unified shape and svbrdf recovery using differentiable monte carlo
  rendering,''
\newblock in {\em Proc. EGSR}, 2021.

\bibitem{Gardner_2019_ICCV}
Marc-Andre Gardner, Yannick Hold-Geoffroy, Kalyan Sunkavalli, Christian Gagne,
  and Jean-Francois Lalonde,
\newblock ``Deep parametric indoor lighting estimation,''
\newblock in {\em Proc. ICCV}, 2019.

\bibitem{deeplightingCVPRW2020}
V.~Gkitsas, N.~Zioulis, F.~Alvarez, D.~Zarpalas, and P.~Daras,
\newblock ``Deep lighting environment map estimation from spherical
  panoramas,''
\newblock in {\em Proc. CVPR Workshop}, 2020.

\bibitem{schoenberger2016sfm}
Johannes~Lutz Sch\"{o}nberger and Jan-Michael Frahm,
\newblock ``Structure-from-motion revisited,''
\newblock in {\em Proc. CVPR}, 2016.

\bibitem{schoenberger2016mvs}
Johannes~Lutz Sch\"{o}nberger, Enliang Zheng, Marc Pollefeys, and Jan-Michael
  Frahm,
\newblock ``Pixelwise view selection for unstructured multi-view stereo,''
\newblock in {\em Proc. ECCV}, 2016.

\bibitem{maurer2018combining}
Daniel Maurer, Yong~Chul Ju, Michael Breu{\ss}, and Andr{\'e}s Bruhn,
\newblock ``Combining shape from shading and stereo: A joint variational method
  for estimating depth, illumination and albedo,''
\newblock {\em IJCV}, 2018.

\bibitem{thies2019deferred}
Justus Thies, Michael Zollh{\"o}fer, and Matthias Nie{\ss}ner,
\newblock ``Deferred neural rendering: Image synthesis using neural textures,''
\newblock in {\em ACM Trans. Graph.}, 2019.

\bibitem{sitzmann2019scene}
Vincent Sitzmann, Michael Zollh{\"o}fer, and Gordon Wetzstein,
\newblock ``Scene representation networks: Continuous 3d-structure-aware neural
  scene representations,''
\newblock in {\em Proc. NeurIPS}, 2019.

\bibitem{mildenhall2020nerf}
Ben Mildenhall, Pratul~P. Srinivasan, Matthew Tancik, Jonathan~T. Barron, Ravi
  Ramamoorthi, and Ren Ng,
\newblock ``Nerf: Representing scenes as neural radiance fields for view
  synthesis,''
\newblock in {\em Proc. ECCV}, 2020.

\bibitem{yu2020pixelnerf}
Alex Yu, Vickie Ye, Matthew Tancik, and Angjoo Kanazawa,
\newblock ``{pixelNeRF}: Neural radiance fields from one or few images,''
\newblock in {\em Proc. CVPR}, 2021.

\bibitem{liu2020neural}
Lingjie Liu, Jiatao Gu, Kyaw~Zaw Lin, Tat-Seng Chua, and Christian Theobalt,
\newblock ``Neural sparse voxel fields,''
\newblock {\em Proc. NeurIPS}, 2020.

\bibitem{barron2021mipnerf}
Jonathan Barron, Ben Mildenhall, Matthew Tancik, Peter Hedman, Ricardo~M. B.,
  and Pratul~P. Srinivasan,
\newblock ``Mip-nerf: A multiscale representation for anti-aliasing neural
  radiance fields,''
\newblock in {\em Proc. ICCV}, 2021.

\bibitem{NimierDavidVicini2019Mitsuba2}
Merlin Nimier-David, Delio Vicini, Tizian Zeltner, and Wenzel Jakob,
\newblock ``Mitsuba 2: A retargetable forward and inverse renderer,''
\newblock {\em ACM Trans. Graph.}, 2019.

\bibitem{RenderingTOG1986}
James~T. Kajiya,
\newblock ``The rendering equation,''
\newblock in {\em Siggraph}, 1986.

\bibitem{BeckmanTOG1982}
R.~L. Cook and K.~E. Torrance,
\newblock ``A reflectance model for computer graphics,''
\newblock {\em ACM Trans. Graph.}, 1982.

\bibitem{GGXEG2007}
Bruce Walter, Stephen~R. Marschner, Hongsong Li, and Kenneth~E. Torrance,
\newblock ``Microfacet models for refraction through rough surfaces,''
\newblock in {\em Eurographics Conference on Rendering Techniques}, 2007.

\bibitem{rother2004grabcut}
Carsten Rother, Vladimir Kolmogorov, and Andrew Blake,
\newblock ``Grabcut: interactive foreground extraction using iterated graph
  cuts,''
\newblock {\em ACM Trans. Graph.}, 2004.

\bibitem{lirsiggraphasia2019}
Rui Li and Wolfgang Heidrich,
\newblock ``Hierarchical and view-invariant light field segmentation by
  maximizing entropy rate on 4d ray graphs,''
\newblock in {\em ACM Trans. Graph.}, 2019.

\bibitem{jensen2014large}
Rasmus Jensen, Anders Dahl, George Vogiatzis, Engil Tola, and Henrik Aan{\ae}s,
\newblock ``Large scale multi-view stereopsis evaluation,''
\newblock in {\em Proc. CVPR}, 2014.

\end{thebibliography}

\end{document}